%% file: main-nv.tex
\pdfoutput=1
\documentclass[10pt,logo,copyright]{nvidiatechreport}
\input{packages}

\definecolor{cvprblue}{rgb}{0.21,0.49,0.74}
\definecolor{srLinkGreen}{HTML}{2F7D6A}
\definecolor{asenashade}{RGB}{237,246,240}
\hypersetup{
  colorlinks=true,
  allcolors=cvprblue,
  pdftitle={ASENA: Self-evolving Agents for Embodied Navigation},
  pdfauthor={An-Chieh Cheng, Isabella Liu, Edmund Bu, Johan Bjorck, Hongxu Yin, Zhengyi Luo, Jan Kautz, Linxi “Jim” Fan, Yuke Zhu, Sifei Liu}
}
\setcitestyle{numbers,square}
\makeatletter
\AtBeginDocument{\immediate\write\@auxout{\string\citation{asenaBibStyle}}}
\makeatother
\newcolumntype{Y}{>{\centering\arraybackslash}X}
\definecolor{darkred}{rgb}{0.7, 0.0, 0.0}

\usepackage{pifont}

\usepackage{wrapfig}

\usepackage[nameinlink]{cleveref}
\crefname{equation}{Eq.}{Eqs.}
\crefname{figure}{Fig.}{Figs.}
\crefname{section}{Sec.}{Sec.}
\crefname{appendix}{App.}{App.}
\crefname{table}{Tab.}{Tabs.}
\usepackage{longtable}
\usepackage{etoolbox}

\input{macros}
\renewcommand{\link}{https://research.nvidia.com/labs/gear/asena}
\newcommand{\asena}{ASENA\xspace}
\newcommand{\capravln}{ASENA-VLN\xspace}
\newcommand{\claudesonnet}{\mbox{\texttt{Claude Sonnet 5}}\xspace}
\newcommand{\gptastra}{\mbox{\texttt{GPT-6 Astra}}\xspace}
\newcommand{\figref}[1]{Fig.~\ref{#1}}
\newcommand{\tabref}[1]{Table~\ref{#1}}
\newcommand{\secref}[1]{Sec.~\ref{#1}}
\title{ASENA: Self-evolving Agents for Embodied Navigation}

\author[1]{An-Chieh Cheng}
\author[1]{Isabella Liu}
\author[2]{Edmund Bu}
\author[1]{Johan Bjorck}
\author[1]{Hongxu Yin}
\author[1]{Zhengyi Luo}
\author[1]{Jan Kautz}
\author[1]{\mbox{Linxi “Jim” Fan}}
\author[1,3]{Yuke Zhu}
\author[1]{Sifei Liu}
\affil[1]{NVIDIA}
\affil[2]{University of California, San Diego}
\affil[3]{The University of Texas at Austin}
\patchcmd{\maketitle}{\vskip20pt}{\vskip6pt}{}{\PackageError{main-nv}{Title spacing patch failed}{Check the report class title definition.}}

\begin{document}
\input{section/icra-nv/00-abstract}
\maketitle

\input{section/icra-nv/teaser}
\par\smallskip
\Needspace{4\baselineskip}
\noindent{\headingfont Abstract}\par\nobreak
{\absfont\noindent\theabstract\par}

\input{section/icra-nv/01-introduction}
\input{section/icra-nv/02-related-work}
\input{section/icra-nv/03-method}

\input{section/icra-nv/04-experiments}
\input{section/icra-nv/05-discussion}

\begingroup
\setcitestyle{numbers}
\bibliographystyle{IEEEtranN}
\bibliography{references-icra-nv,references-supplement-nv}
\endgroup

\appendix
\newpage
\clearpage
\input{section/07-supplement-nv}

\end{document}

%% file: packages.tex
\usepackage[authoryear,sort&compress,round]{natbib}

\usepackage[utf8]{inputenc} %
\usepackage[T1]{fontenc}    %

\usepackage{parskip}        %
\usepackage{url}            %
\usepackage{booktabs}       %
\usepackage{amsfonts}       %
\usepackage{nicefrac}       %
\usepackage{microtype}      %
\usepackage{xcolor}         %
\usepackage[dvipsnames]{xcolor} %
\usepackage{graphicx}
\usepackage{animate}        %
\usepackage{subcaption}
\usepackage{tabularx}
\usepackage{makecell}
\usepackage{adjustbox}
\usepackage{setspace}
\newcolumntype{M}[1]{>{\centering\arraybackslash}m{#1}}
\usepackage{float}
\usepackage{tikz}
\usetikzlibrary{positioning,shapes,arrows}
\usepackage{amsmath,amsfonts,bm, bbm}
\usepackage{multirow}
\usepackage{comment}
\usepackage{gensymb}
\usepackage{lipsum}
\usetikzlibrary{arrows.meta, positioning, fit}
\usepackage[para]{threeparttable}
\usepackage{tikz}
\usetikzlibrary{tikzmark}
\usepackage{hyperref}       %

%% file: macros.tex
\usepackage{mathtools}
\usepackage{enumitem}
\usepackage{caption}

\renewcommand{\paragraph}[1]{{\vspace{1mm}\noindent \bf #1}.}
\newcommand{\link}{https://asena-bot.github.io/}

%% file: section/icra-nv/00-abstract.tex
\begin{abstract}
We present \asena, an embodied agent system that enables general-purpose coding agents to program humanoid navigation, manipulation, and interaction. Through a unified robot harness, agents read sensor data, write and execute programs, inspect recorded outcomes, repair failures, and reuse notes and executable skills while keeping their model weights fixed. We also introduce \capravln, a 4B monocular navigation policy that serves as an optional tool within this programmable system. \capravln predicts body-frame trajectories for both extended routes and short-horizon behaviors. Its shared vision-language decoder is trained on route instructions, visual question answering, and a newly curated dataset of geometry-derived atomic navigation tasks. As a standalone policy, \capravln achieves state-of-the-art success rates of 68.7\% on R2R and 70.2\% on RxR. When integrated with a coding agent, learned navigation improves \asena's success rate by 11 percentage points on both agentic benchmarks and reduces execution time. Persistent workspace evolution uses simulator feedback to improve success over ten passes through recurring 100-task subsets. Success increases from 72\% to 98\% on R2R and from 65\% to 89\% on RxR. On embodied question answering, \asena achieves state-of-the-art accuracy with fewer interaction steps. On a Unitree G1, we demonstrate search, visual inspection, spatial reasoning, synthesized gestures, and mobile manipulation combining object pickup, carrying, and handoff without a pre-built map. These demonstrations illustrate how online programming extends robot behavior beyond route following and predefined skills.
\end{abstract}

%% file: section/icra-nv/teaser.tex
\begin{minipage}{\textwidth}
  \normalfont
  \captionsetup{hypcap=false}
  \makeatletter
  \def\@captype{figure}
  \makeatother
  \centering
  \includegraphics[width=\textwidth,trim=3bp 0bp 3bp 0bp,clip]{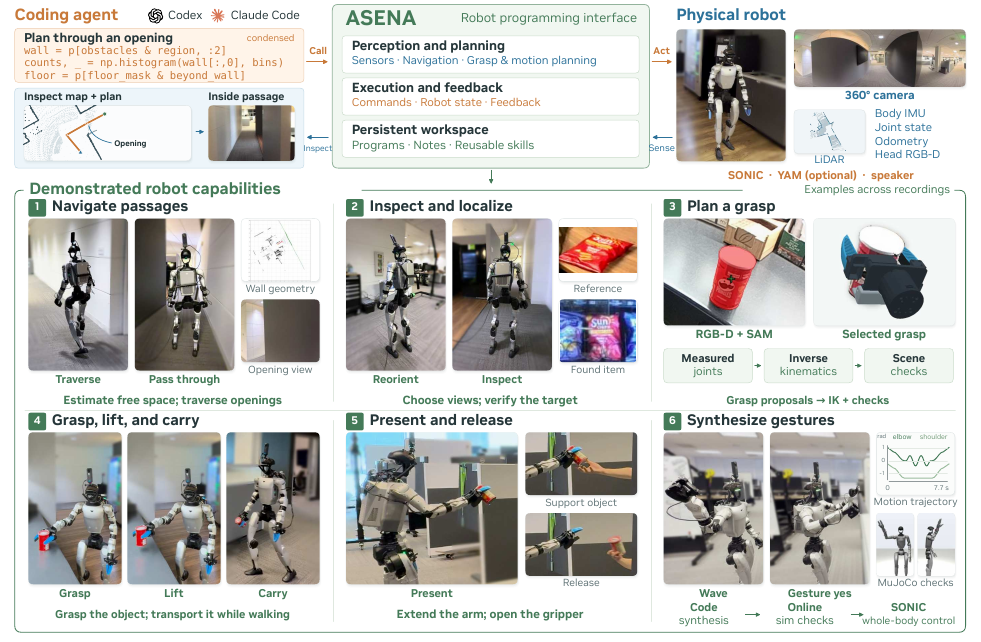}
\caption{\textbf{\asena connects coding agents to physical robot skills.}
Programs combine sensing, computation, robot execution, and reusable
skills, with learned navigation as an optional tool. The coding example uses
LiDAR geometry to propose a passage route. The gallery illustrates navigation,
inspection, grasp planning, object transport, handoff, and gestures across
recorded demonstrations. \href{\link}{\nolinkurl{https://research.nvidia.com/labs/gear/asena}}.}

  \label{fig:teaser}
\end{minipage}

%% file: section/icra-nv/01-introduction.tex
\section{Introduction}
A useful humanoid robot must do more than reach a destination. If asked to
check whether a vending machine contains an item and return with an answer, it
must decide where to search, recognize the item, remember the user's location,
and communicate what it found. Retrieving an object from a room also requires
planning a grasp, picking it up, carrying it back, and handing it to the user.
Navigation is therefore only one part of a larger task whose next step depends
on what the robot observes.

A learned navigation policy provides movement capabilities, but such requests
also require deciding what to inspect, how to act on objects, and what to
communicate. These procedures depend on new observations and may extend beyond
predefined skills.
Coding agents can construct and revise them during a mission through
programmatic access to sensing, computation, simulation, and action. Physical
deployment also requires feedback tied to outcomes and a persistent workspace for retaining useful
procedures.

We present \asena, an embodied agent system that gives coding agents
programmatic control of humanoid tasks through a unified robot harness
(\figref{fig:teaser}). The agent can read sensors, derive geometry, inspect
visualizations, and write programs that combine navigation, grasp planning,
whole-body motion, and user interaction. Linked execution records let the agent
inspect outcomes and revise these programs. A persistent workspace
stores notes and executable skills. The system can improve through program
revision while keeping its model weights fixed.

Learned policies can further expand this programmable toolset when suitable
training data are available. Delegating motion to a specialized policy reduces
repeated agent decisions and supports navigation behaviors that are difficult
to specify geometrically. We develop \capravln, a 4B monocular navigation
policy trained on route instructions and geometry-derived navigation atoms.
Building on vision-and-language navigation
(VLN)~\cite{anderson2018r2r,ku2020rxr,zhang2024navid,cheng2025navila,wei2025streamvln},
\capravln predicts body-frame trajectories for both extended routes and
short-horizon goals, such as approaching an object or passing through a
doorway. We curate a new atomic navigation dataset that directly supervises
these reusable, fine-grained behaviors and complements conventional
route-level training. The policy remains an optional tool: a program may call
it directly, combine it with geometric navigation, or complete a task using
other available operations.

We make three contributions. First, we introduce a robot harness that lets
coding agents program humanoid navigation, manipulation, and interaction while
retaining executable skills and experience. Second, we develop a state-of-the-art
monocular navigation policy that supports both route following and reusable
atomic skills. Third,
we show through evaluation that workspace evolution improves task performance
while model weights remain fixed. We evaluate
navigation and embodied question answering (EQA) in simulation and demonstrate
navigation, mobile manipulation, and user interaction on a real-world Unitree G1
robot operating without a pre-built map.

%% file: section/icra-nv/02-related-work.tex
\section{Related Work}
\paragraph{Code as policies}
Code as Policies~\cite{liang2022codeaspolicies} composes perception and control
APIs in programs. CaP-X~\cite{fu2026capx} provides CaP-Gym and benchmarks how
API abstraction and execution feedback affect coding agents for manipulation.
ASPIRE~\cite{lu2026aspire} uses multimodal execution traces to diagnose failures,
repair robot programs, and retain reusable skills.

\paragraph{Agentic self-improvement}
Voyager~\cite{wang2023voyager} retains and repairs executable skills.
Reflexion~\cite{shinn2023reflexion} retains textual feedback without weight
updates. ENPIRE~\cite{xiao2026enpire} couples real-robot reset and verification
with agent-driven program synthesis and policy training. \asena studies how a
persistent workspace can evolve around a pretrained navigation policy while
all model weights remain fixed. We quantify this workspace evolution in
simulation and demonstrate the system's capabilities on a physical
robot.

\paragraph{Language-guided navigation}
Topological approaches such as DUET and ETPNav~\cite{chen2022duet,an2023etpnav}
address visual instruction following. Video-to-action policies such as NaVid,
NaVILA, and StreamVLN~\cite{zhang2024navid,cheng2025navila,wei2025streamvln}
address the same task. LM-Nav~\cite{shah2023lm} composes pretrained models.
InstructNav~\cite{long2024instructnav} combines zero-shot reasoning with generic
motion primitives. DualVLN and Qwen-RobotNav~\cite{wei2025dualvln,qwen2026robotnav}
integrate high-level reasoning with learned navigation policies.
Uni-LaViRA~\cite{ding2026unilavira} studies unified language-vision-action
translation. Our VLN policy is most closely related to video-to-action
navigation policies. We focus on a lightweight 4B model with fine-grained
atomic capabilities that coding agents can compose with other tools and
retained programs. We evaluate both its standalone navigation performance and
its utility within the full embodied agent system.

\paragraph{Embodied question answering}
Explore-EQA~\cite{ren2024exploreeqa} studies active exploration until an agent
can answer with sufficient confidence. 3D-Mem and
GraphEQA~\cite{yang2025threedmem,saxena2025grapheqa} organize observations into
spatial memories. MemoryEQA~\cite{zhai2025memoryeqa} extends the setting to
longer-horizon, multi-target questions. FAST-EQA~\cite{zhang2026fasteqa} focuses
on efficient evidence gathering. We use these EQA settings to evaluate
\asena's ability to compose perception, navigation, and memory tools. We also
evaluate its ability to improve by revising persistent notes and executable skills
while keeping model weights fixed.

%% file: section/icra-nv/03-method.tex
\section{Method}
\label{sec:method}

\begin{figure*}[t]
  \centering
  \includegraphics[width=\textwidth]{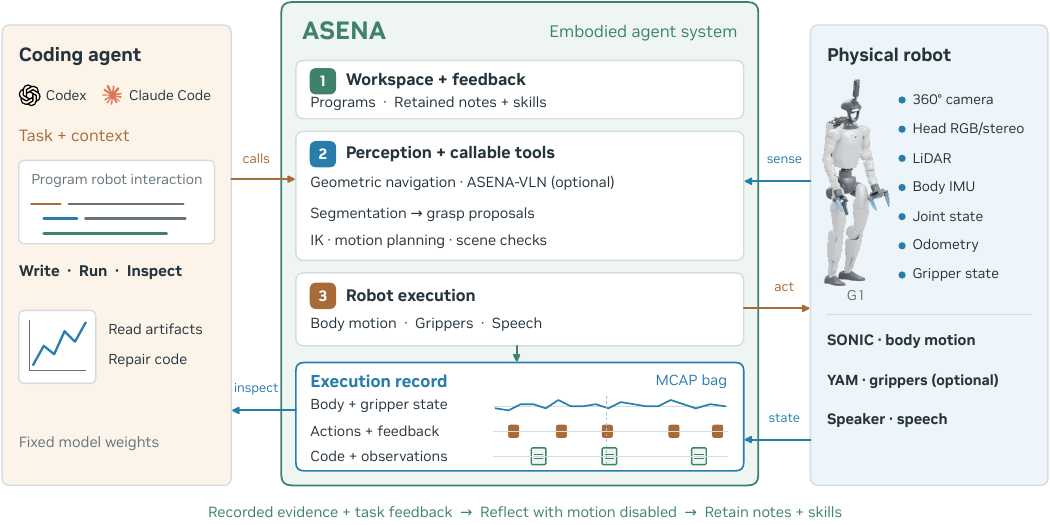}
\caption{\textbf{\asena connects coding agents to physical robots.}
The system integrates robot sensing, a persistent workspace, navigation and
manipulation tools, and robot execution. Body motion uses SONIC. Optional
YAM grippers use a separate controller. The system stores each run as
an MCAP record~\cite{foxglove2024mcap} that aligns sensor and robot states
with actions, feedback, code, and visualizations for inspection and
replay. The agent reflects on these records to update its notes and skills.}
  \label{fig:overview}
\end{figure*}

\asena gives a coding agent programmatic access to a physical robot
(\figref{fig:overview}). It brings sensing, computation, visualization,
simulation, and action into a unified workflow for writing, running, and
inspecting programs. The system records each run for later inspection, repair,
and reuse. The agent can also invoke
a learned navigation policy as an optional tool within the same interface.

\input{figures/icra-nv/policy-architecture}

\subsection{Grounding Programs in Robot Feedback}

On G1, \asena exposes a 360-degree camera, LiDAR, odometry, a body IMU, and joint
states, with a calibrated head RGB/stereo camera for manipulation.
The coding agent processes and visualizes these inputs to estimate
geometry, inspect objects, and monitor execution. When an action differs from
its intent, the agent can inspect the recorded evidence and revise its program.

\paragraph{Navigation}
The agent can plan motion with geometric tools or invoke \capravln for
language-conditioned navigation. Geometric tools use LiDAR-based occupancy
maps to plan waypoints and check obstacle clearance. \capravln predicts bounded
body-frame trajectories from language and recent monocular observations
(\secref{sec:navtool}). The agent chooses when to invoke each tool, checks
progress through new observations, and revises the plan as needed.

\paragraph{Manipulation}
SAM~2~\cite{ravi2024sam2} object masks and depth measurements
ground candidate grasps.
The system can supplement geometric proposals with learned candidates from
GraspGenX~\cite{han2026graspgenx} when additional grasp options are needed.
Inverse kinematics and motion planning convert these candidates into
trajectories for reaching, lifting, and carrying. Before execution, the system
rebuilds arm trajectories from measured joint states and checks joint limits,
sampled self-collisions, and scene clearance. These checks include held-object
clearance during lift and retraction.
A separate controller opens and closes the optional YAM parallel grippers and
reports jaw width and controller state. The agent can compose these tools with
navigation to program object pickup and handoff.

\begin{samepage}
\paragraph{Skill composition and execution}
\asena exposes sensing, control, and safety operations as callable atomic
skills. The coding agent combines navigation, grasp planning, arm trajectories,
and gripper commands with computation and interaction logic to create reusable
task programs. It writes and revises task logic and motion plans using execution
feedback, while physical execution interfaces remain fixed. The SONIC whole-body
controller~\cite{luo2025sonic} executes body motion. Gripper commands use the
separate YAM controller, and speech uses the onboard speaker.

\end{samepage}

\paragraph{Execution checks}
\asena validates generated gestures online in MuJoCo for pose continuity,
joint and motion limits, and sampled self-collision. Failed validation returns
the violated constraint to the agent for repair. Physical motion requires
validation and operator approval, including motion requested by revised
programs. An independent navigation safeguard enforces motion limits and
rejects commands when odometry is stale or the battery level is low.
Heartbeat-triggered stops and manual emergency halts provide additional
protection.

\input{tables/icra-nv/vln-policy}
\input{tables/icra-nv/objectnav}
\input{tables/icra-nv/navigation}

\subsection{Turning Execution into Retained Experience}

\asena records runs in MCAP format~\cite{foxglove2024mcap}. These records align
sensor and robot states with actions, feedback, code, and visualizations.
Replaying this evidence lets the agent identify where execution diverged from
its intent and repair the responsible program.

After repairing a failure, \asena can retain the solution for future tasks.
We represent the workspace at revision $k$ as $W_k=(N_k,S_k)$, where $N_k$
contains notes and $S_k$ contains executable skills. A fixed coding model
consults this workspace while solving tasks. Between revisions, a separate
fixed-weight improver uses execution traces $\mathcal{T}_k$ and feedback
$f_k$ to update the workspace:
\begin{equation}
  W_{k+1}=I_\phi(W_k,\mathcal{T}_k,f_k).
  \label{eq:revision}
\end{equation}

Model and controller weights remain fixed. In simulation, private workers
may edit workspaces between passes, while writes are disabled during frozen
evaluation. On G1, execution records and optional operator feedback support
skill revision while robot motion is disabled.

\subsection{Learning a Reusable Visual Navigation Policy}
\label{sec:navtool}

\capravln reduces repeated agent reasoning on challenging routes. This
optional tool predicts bounded body-frame trajectories from language and
recent monocular observations. The coding agent decides when to invoke it,
checks progress, and combines it with other tools. Policy training expands
navigation capabilities within the same programming interface.

\paragraph{Policy inputs}
We extend a lightweight Qwen3-VL-4B-Instruct~\cite{qwen2025qwen3vl} with
dedicated token vocabularies for trajectories and pixel goals
(\figref{fig:policy-architecture}). The model takes up to eight front-view
frames and a structured JSON instruction. This interface makes the policy easy
to invoke as a callable tool. The instruction specifies a concise navigation
command and can optionally include a desired end pose, anchor objects, or motion
constraints. Recency-weighted visual tokens preserve temporal context while
emphasizing the latest observation.

\paragraph{Output representations}
A shared autoregressive decoder predicts up to eight cumulative body-frame
waypoints, each represented as $(x,y,\theta)$. Fixed bins along each axis convert
trajectory tokens into explicit motion targets without requiring a separate
continuous action head. Pixel-goal tokens provide auxiliary supervision by
encouraging the model to localize its intended destination in the image.
Visual question answering retains the model's original text vocabulary.
A pixel goal is not directly executable and requires an additional motion
planner or, in simulation, a shortest-path follower. All evaluations in this
paper therefore use trajectory-token predictions rather than pixel goals.

\paragraph{Curated atomic-navigation data}
A callable navigation policy must follow both extended routes and precise
local commands. We therefore curate nine atomic navigation tasks in MP3D and
HM3D. They cover behaviors such as moving a specified distance, rotating to a target heading, approaching an object from a given direction, aligning with a
target, entering a room, passing through a doorway, following a corridor, and moving between floors. We generate and validate each task using geodesic
distance, visibility, and scene geometry so that every instruction describes
a feasible and measurable motion. Each instruction is paired with a waypoint
trajectory and auxiliary pixel-goal supervision. These data directly teach
the reusable, fine-grained operations that task programs can invoke and
compose.

\paragraph{Training mixture}
We train \capravln on a mixture of these atomic tasks, short-horizon ObjectNav,
and vision-language-navigation data from R2R, RxR, ScaleVLN, and
SRDF~\cite{anderson2018r2r,ku2020rxr,wang2023scalevln,wang2025srdf}. We also
include multi-view VQA data~\cite{sensenova-si} to strengthen the model's spatial
understanding. To match our callable policy interface, we use
GPT-5.5~\cite{openai2026gpt55} to re-caption the original navigation instructions
into our structured JSON format. The re-captioned instructions include concise
navigation commands and, when applicable, end poses, anchor objects, and motion
constraints. The reported checkpoint uses fewer than 10K optimizer steps and
fewer than 5M training samples, including repeats. Training uses a 3:1 mixture
of trajectory prediction and spatial VQA.

\input{figures/icra-nv/navigation-evolution}

%% file: figures/icra-nv/policy-architecture.tex
\begin{figure}[t]
  \centering
  \includegraphics[width=\textwidth]{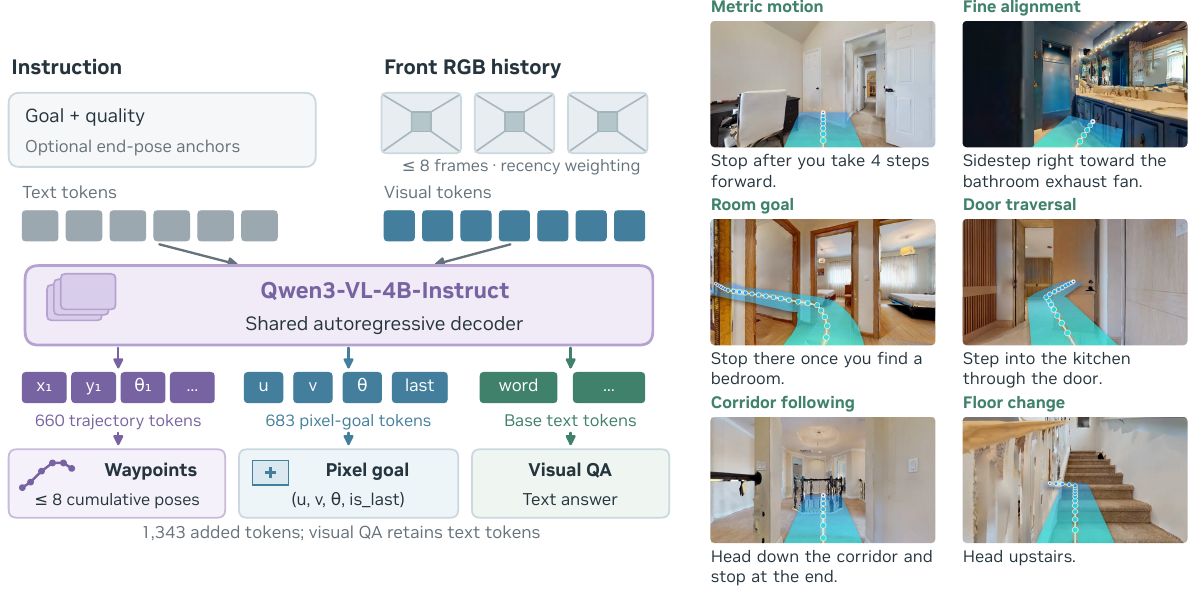}
  \caption{\textbf{Navigation policy architecture and atomic training examples.}
  Left: language and monocular
  history enter a shared autoregressive decoder. Separate token blocks encode
  trajectories and pixel goals. Visual QA uses text tokens. Motion tokens decode
  to body-frame waypoints.
  Architecture glyphs are schematic. Right: recorded training observations with
  instructions and overlaid trajectories.
  We show more examples in \Cref{fig:atomic-training-gallery}.}
  \label{fig:policy-architecture}
\end{figure}

%% file: tables/icra-nv/vln-policy.tex
\begin{table*}[t]
  \centering
  \caption{Standalone monocular VLN-CE policy results on R2R~\cite{anderson2018r2r} and RxR~\cite{ku2020rxr} \texttt{val\_unseen}. ASENA-VLN-4B achieves a state-of-the-art result compared to prior work, with significantly lower navigation error and higher nDTW.}
  \label{tab:vln-policy}
  \footnotesize
  \setlength{\tabcolsep}{6pt}
  \renewcommand{\arraystretch}{1.02}
  \begin{tabularx}{\textwidth}{l*{8}{Y}}
    \toprule
    & \multicolumn{4}{c}{R2R} & \multicolumn{4}{c}{RxR} \\
    \cmidrule(lr){2-5}\cmidrule(lr){6-9}
    Method & NE $\downarrow$ & OSR $\uparrow$ & SR $\uparrow$ & SPL $\uparrow$
           & NE $\downarrow$ & nDTW $\uparrow$ & SR $\uparrow$ & SPL $\uparrow$ \\
    \midrule
  NaVid~\cite{zhang2024navid} & 5.72 & 49.2 & 41.9 & 36.5 & 5.72 & -- & 45.7 & 38.2 \\
  Uni-NaVid~\cite{zhang2025uninavid} & 5.58 & 53.3 & 47.0 & 42.7 & 6.24 & -- & 48.7 & 40.9 \\
  NaVILA~\cite{cheng2025navila} & 5.22 & 62.5 & 54.0 & 49.0 & 6.77 & 58.8 & 49.3 & 44.0 \\
  StreamVLN~\cite{wei2025streamvln} & 4.98 & 64.2 & 56.9 & 51.9 & 6.22 & 61.9 & 52.9 & 46.0 \\
  NaVIDA~\cite{zhu2026navida} & 4.32 & 69.5 & 61.4 & 54.7 & 5.23 & 67.0 & 57.4 & 49.6 \\
  JanusVLN~\cite{zeng2026janusvln} & 4.78 & 65.2 & 60.5 & 56.8 & 6.06 & 62.1 & 56.2 & 47.5 \\
  DecoVLN~\cite{xin2026decovln} & 5.01 & 63.5 & 56.3 & 50.5 & 5.73 & 63.5 & 54.2 & 46.3 \\
  ActiveVLN~\cite{zhang2026activevln} & 5.31 & 58.3 & 50.1 & 43.7 & 5.84 & 58.1 & 50.7 & 41.2 \\
  Qwen-VLA-Instruct~\cite{wang2026qwenvla} & 5.10 & 69.0 & 57.5 & 51.2 & 5.80 & 57.1 & 59.6 & 47.8 \\
  InternVLA-N1 (S2, SPF)~\cite{cai2025internvlan1} & 4.25 & 68.3 & 60.9 & 55.2 & 5.71 & 46.8 & 63.5 & 55.0 \\
  RynnBrain-Nav-8B (SPF)~\cite{dang2026rynnbrain} & 4.92 & 71.6 & 58.6 & 49.6 & 6.20 & 59.6 & 56.1 & 49.6 \\
  DualVLN~\cite{wei2025dualvln} & \underline{4.05} & 70.7 & 64.3 & 58.5 & 4.58 & \underline{70.0} & 61.4 & 51.8 \\
  InternVLA-N1~\cite{cai2025internvlan1} & 4.83 & 63.3 & 58.2 & 54.0 & 5.91 & 65.3 & 53.5 & 46.1 \\
  Qwen-RobotNav-4B~\cite{qwen2026robotnav} & 4.22 & \textbf{73.6} & \underline{66.9} & \underline{60.5} & \underline{4.15} & 68.6 & \textbf{71.3} & \textbf{61.5} \\
    \midrule
  \rowcolor{asenashade}
  ASENA-VLN-4B & \textbf{3.52} & \underline{72.6} & \textbf{68.7} & \textbf{64.2} & \textbf{3.90} & \textbf{73.1} & \underline{70.2} & \underline{59.7} \\
    \bottomrule
  \end{tabularx}
\end{table*}

%% file: tables/icra-nv/objectnav.tex
\begin{table}[t]
  \centering
  \caption{Object-goal navigation: paired SR/SPL (\%, both $\uparrow$).
  Published baseline values are from Qwen-RobotNav~\cite{qwen2026robotnav}
  and SysNav~\cite{zhu2026sysnav}.
  $^\ddagger$ denotes HM3D v1 result, where emphasis excludes these.}
  \label{tab:objectnav}
  \footnotesize
  \setlength{\tabcolsep}{2pt}
  \renewcommand{\arraystretch}{1.04}
  \begin{tabularx}{\columnwidth}{l*{4}{Y}}
    \toprule
    & HM3D & \multicolumn{3}{c}{HM3D-OVON} \\
    \cmidrule(lr){3-5}
    Method & v2 val & seen & syn. & unseen \\
    \midrule
    VLFM~\cite{yokoyama2024vlfm} & 63.6/32.5 & 35.2/18.6 & 32.4/17.3 & 35.2/19.6 \\
    OpenFMNav$^{\ddagger}$~\cite{kuang2024openfmnav} & 52.5/24.1 & -- & -- & -- \\
    SG-Nav~\cite{yin2024sgnav} & 49.6/25.5 & -- & -- & -- \\
    TriHelper$^{\ddagger}$~\cite{zhang2024trihelper} & 56.5/25.3 & -- & -- & -- \\
    WMNav$^{\ddagger}$~\cite{nie2025wmnav} & 58.1/31.2 & -- & -- & -- \\
    CogNav$^{\ddagger}$~\cite{cao2024cognav} & 72.5/26.2 & -- & -- & -- \\
    Uni-NaVid$^{\ddagger}$~\cite{zhang2025uninavid} & 73.7/37.1 & 41.3/21.1 & 43.9/21.8 & 39.5/19.8 \\
    DAgRL+OD~\cite{yokoyama2024ovon} & -- & 38.5/21.1 & 39.0/21.4 & 37.1/19.8 \\
    MTU3D~\cite{zhu2025mtu3d} & -- & 55.0/23.6 & 45.0/14.7 & 40.8/12.1 \\
    NavFoM~\cite{zhang2025navfom} & -- & 40.1/27.1 & 45.4/32.6 & 45.2/\underline{31.9} \\
    ABot-N0~\cite{chu2026abotn0} & -- & 55.3/\underline{32.1} & 55.4/\underline{33.2} & \underline{54.0}/30.5 \\
    ApexNav~\cite{zhang2025apexnav} & \underline{76.2}/\underline{38.0} & -- & -- & -- \\
    Qwen-RobotNav-4B~\cite{qwen2026robotnav} & 75.6/30.6 & \underline{57.7}/24.4 & \underline{60.1}/25.1 & 53.1/20.9 \\
    \midrule
    \rowcolor{asenashade}
    \asena & \textbf{80.8}/\textbf{38.1} & \textbf{60.5}/\textbf{33.5} & \textbf{65.8}/\textbf{37.9} & \textbf{64.9}/\textbf{38.4} \\
    \bottomrule
  \end{tabularx}
  \par\smallskip
\end{table}

%% file: tables/icra-nv/navigation.tex
\begin{table*}[t]
\centering
\caption{Navigation before self-evolution on R2R and RxR Agentic Splits. NE is in meters and OSR/SR/SPL are percentages. Calls/ep. is mean tool calls over 100 episodes. Green percentages show reductions from the same backend without VLN.}
\label{tab:navigation}
\footnotesize
\setlength{\tabcolsep}{1.8pt}
\definecolor{callgain}{RGB}{25,112,78}
\newcommand{\callreduction}[1]{{\scriptsize\textcolor{callgain}{$\downarrow$\,#1\%}}}
\begin{tabularx}{\textwidth}{lc*{4}{Y}rl*{3}{Y}rl}
\toprule
 & & \multicolumn{6}{c}{R2R Agentic Split} & \multicolumn{5}{c}{RxR Agentic Split} \\
\cmidrule(lr){3-8}\cmidrule(lr){9-13}
Method / coding agent & VLN tool & NE $\downarrow$ & OSR $\uparrow$ & SR $\uparrow$ & SPL $\uparrow$ & \multicolumn{2}{c}{Calls/ep. $\downarrow$} & NE $\downarrow$ & SR $\uparrow$ & SPL $\uparrow$ & \multicolumn{2}{c}{Calls/ep. $\downarrow$} \\
\midrule
\multicolumn{13}{l}{\emph{Published zero-shot methods}} \\
InstructNav~\cite{long2024instructnav} & -- & 6.89 & 47.0 & 31.0 & 24.0 & -- & & -- & -- & -- & -- & \\
Open-Nav~\cite{qiao2025opennav} & -- & 6.70 & 23.0 & 19.0 & 16.1 & -- & & -- & -- & -- & -- & \\
CA-Nav~\cite{chen2024canav} & -- & 7.58 & 48.0 & 25.3 & 10.8 & -- & & 10.4 & 19.0 & 6.0 & -- & \\
GC-VLN~\cite{yin2025gcvln} & -- & 7.30 & 41.8 & 33.6 & 16.3 & -- & & 8.80 & 33.8 & 13.8 & -- & \\
Three-Step Nav~\cite{zheng2026threestepnav} & -- & 5.87 & 39.0 & 34.0 & 29.1 & -- & & 9.21 & 22.0 & 16.1 & -- & \\
Uni-LaViRA~\cite{ding2026unilavira} & -- & 3.66 & 73.7 & 60.7 & 47.7 & -- & & 6.48 & 51.3 & 34.0 & -- & \\
Minimal (\claudesonnet, SDK)~\cite{zhou2026embodied} & -- & 5.80 & 61.3 & 51.3 & 37.8 & -- & & -- & -- & -- & -- & \\
\midrule
\multicolumn{13}{l}{\emph{\asena policy and fresh-workspace agents}} \\
\rowcolor{asenashade}
ASENA-VLN-4B & -- & 3.16 & 82.0 & 78.0 & 70.7 & -- & & 4.54 & 62.0 & 52.9 & -- & \\
\specialrule{0.3pt}{0pt}{0pt}
\rowcolor{asenashade}
\asena (\claudesonnet) & $\times$ & \underline{5.23} & \underline{70.0} & \underline{61.0} & \underline{34.2} & \underline{83.32} & & \underline{5.80} & \underline{50.0} & \underline{31.9} & \underline{103.83} & \\
\rowcolor{asenashade}
\asena (\claudesonnet) & $\checkmark$ & \textbf{3.31} & \textbf{83.0} & \textbf{72.0} & \textbf{54.9} & \textbf{33.96} & \callreduction{59.2} & \textbf{4.94} & \textbf{61.0} & \textbf{45.5} & \textbf{38.91} & \callreduction{62.5} \\
\specialrule{0.3pt}{0pt}{0pt}
\rowcolor{asenashade}
\asena (\gptastra) & $\times$ & \textbf{1.84} & \textbf{90.0} & \underline{87.0} & \underline{72.4} & \underline{26.37} & & \textbf{0.82} & \textbf{93.0} & \textbf{78.0} & \underline{40.80} & \\
\rowcolor{asenashade}
\asena (\gptastra) & $\checkmark$ & \underline{2.18} & \underline{89.0} & \textbf{88.0} & \textbf{72.8} & \textbf{14.00} & \callreduction{46.9} & \underline{1.65} & \underline{86.0} & \underline{64.0} & \textbf{19.98} & \callreduction{51.0} \\
\bottomrule
\end{tabularx}
\end{table*}

%% file: figures/icra-nv/navigation-evolution.tex
\begin{figure*}[t]
  \centering
  \includegraphics[width=\textwidth]{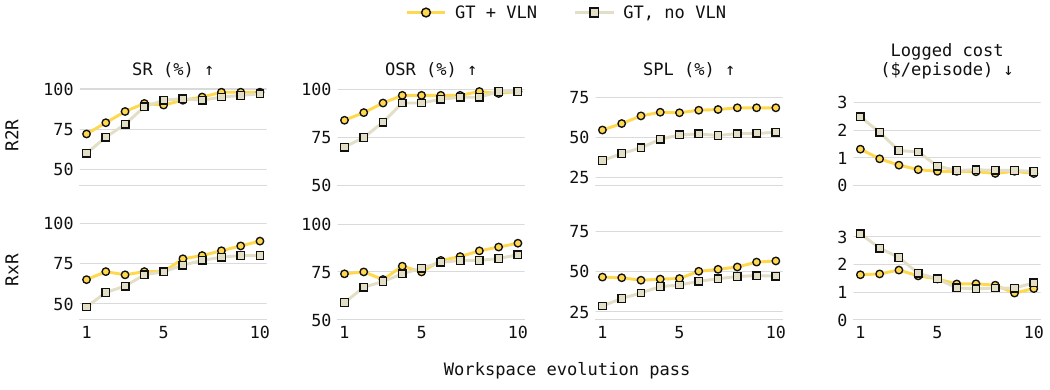}
\caption{\textbf{Navigation workspace self-evolution on Agentic Splits.}
We report results on our R2R and RxR Agentic Splits (100 tasks each), following the
reduced-set evaluation practice of Open-Nav and
Uni-LaViRA~\cite{qiao2025opennav,ding2026unilavira}.
Each pass revisits the same tasks with ground-truth feedback. Workspace
experience is retained while all model weights remain fixed. Success-related
metrics improve and logged cost per episode generally decreases. The VLN tool
consistently improves SPL.}
  \label{fig:navevolve}
\end{figure*}

%% file: section/icra-nv/04-experiments.tex
\section{Evaluation and Applications}
\label{sec:experiments}

We first evaluate \capravln as a standalone instruction-following policy and
then evaluate the full \asena system on object-goal navigation. We next study
how the learned policy and retained workspace experience affect coding-agent
performance with model weights fixed. Finally, we evaluate embodied question
answering in simulation and demonstrate the system on a Unitree G1 robot.

\subsection{Standalone VLN Policy}
\label{sec:standalone}

\tabref{tab:vln-policy} evaluates \capravln independently of coding-agent
planning on R2R's shorter instructions and RxR's longer, more detailed routes.
We use 0.25\,m forward steps and 15-degree rotations. Navigation error (NE)
is the final goal distance in meters. Oracle success rate (OSR) measures
entering the goal region, while success rate (SR) requires stopping there.
SPL accounts for path efficiency and nDTW for reference-route similarity.
All four are percentages. SPF denotes the simulator shortest-path follower
used by pixel-goal methods.

\subsection{Object-Goal Navigation}
\label{sec:objectnav}

\tabref{tab:objectnav} evaluates search for an object that may not initially be
visible. \capravln is trained primarily for atomic navigation behaviors rather
than global exploration. We pair it with a separate
VLM~\cite{google2026gemini31flashlite} that plans where to search and updates the
plan from new observations.

On HM3D-v2 and HM3D-OVON, this combination of global search reasoning and
learned local navigation achieves
state-of-the-art results compared with published methods
from~\cite{qwen2026robotnav,zhu2026sysnav}. Performance remains consistent
across seen, synonymous, and unseen OVON object categories, indicating that
the system can handle variations in both object descriptions and the routes
required to find them.

\subsection{Complementary Agent and Policy Capabilities}
\label{sec:fresh}

\tabref{tab:navigation} studies how coding agents and \capravln contribute to
navigation performance. We compare published zero-shot methods, standalone
\capravln, and \asena instantiated with two coding backends. Our R2R and RxR
\emph{Agentic Splits} each contain 100 tasks, following the reduced-set
evaluation practice of Uni-LaViRA and
Open-Nav~\cite{ding2026unilavira,qiao2025opennav}. All policy and agent
variants use the same task IDs across ten scenes in each benchmark. Four RxR
episodes produce non-finite NE or SPL values. We count them as failures for SR
and compute NE and SPL over the remaining 96 episodes.
Each coding agent starts with a fresh workspace. \claudesonnet runs at high
effort, while \gptastra runs at maximum effort.

\input{tables/icra-nv/eqa}
\input{tables/icra-nv/eqa-workspace}

\paragraph{Strong coding agents navigate effectively without a learned policy}
Using only geometric and programmatic tools, \gptastra achieves
state-of-the-art performance and surpasses standalone \capravln. On R2R, its
success rate also approaches the reported human performance of 90\% SR. Its
strong reasoning allows it to interpret instructions, construct routes,
monitor progress, and recover from errors without relying on a learned
navigation policy. These results establish coding agents themselves as strong
zero-shot navigation systems.

\paragraph{Learned navigation improves weaker coding backends}
The benefit of \capravln is more pronounced for \claudesonnet. Adding the
policy improves both success and path efficiency, increasing SR by 11
percentage points on both R2R and RxR. The specialized policy therefore
compensates for limitations in route execution and enables a weaker coding
backend to complete more tasks. In comparison, \gptastra already performs
strongly without the policy and benefits less from it. On RxR, enabling the
policy slightly reduces its success rate. The value of learned navigation
therefore depends on the coding backend and how effectively it incorporates
policy predictions into its plans.

\paragraph{Learned navigation reduces repeated agent interaction}
A single \capravln call executes several motion decisions, reducing repeated
observation and action requests. Mean tool calls fall for every backend and
benchmark (\tabref{tab:navigation}): by 46.9\%/51.0\% on R2R/RxR for
\gptastra and 59.2\%/62.5\% for \claudesonnet. The policy therefore absorbs
low-level navigation even when success changes little. These call reductions
show that learned navigation can reduce agent interaction while improving
less capable backends, complementing strong agents' direct programmatic control.

\subsection{Retained Experience Improves Recurring Tasks}
\label{sec:navevolve}

We evaluate workspace evolution by repeatedly presenting the same tasks,
scenes, instructions, and initial poses. In each pass, sixteen
\claudesonnet workers at high effort execute the tasks and may retain programs,
notes, and prior action sequences in private workspaces. Between passes, an
improver uses their traces and feedback to update the canonical workspace. We
run campaigns with and without \capravln on R2R and RxR using ground-truth
feedback, while keeping all model, policy, and controller weights fixed.

As shown in \figref{fig:navevolve}, success improves across all four
campaigns. Logged worker and improver costs decrease as the workspace evolves.
These results show that retained experience can improve recurring-task
performance and reduce cost without updating model weights.

The variants with and without \capravln reach similar final success rates on
R2R, while the policy retains a clearer advantage on the longer RxR
instructions. Tool-equipped workspaces also maintain higher SPL on both
benchmarks, showing that \capravln improves path efficiency even when final
success rates converge.

\begin{figure*}[!t]
  \centering
  \includegraphics[width=\textwidth]{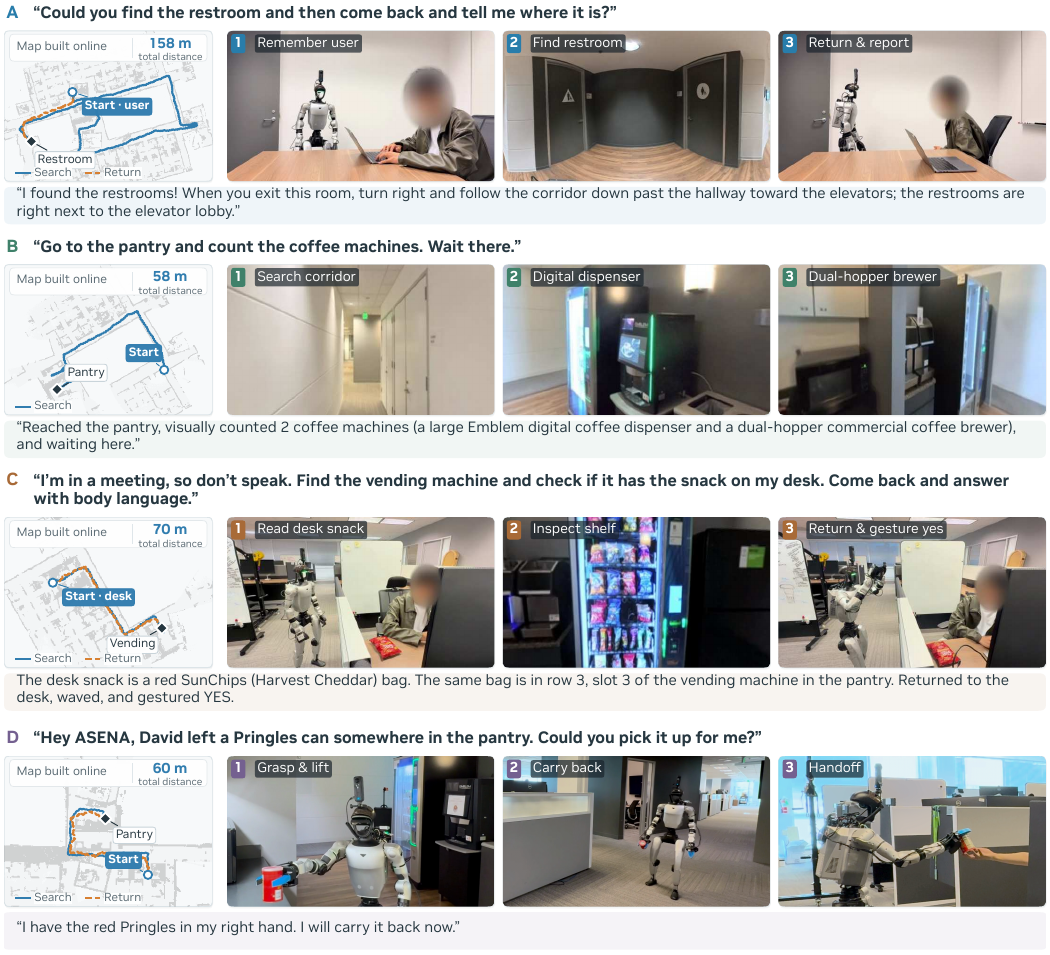}
  \caption{\textbf{Real-world navigation, manipulation, and interaction on the G1 robot.}
  Four demonstrations pair online maps and recorded trajectories
  with observations and robot actions. Blue and dashed orange traces show
  search and return. Distances are estimated from drive odometry.
  D combines pickup, carrying, and handoff. Prompts and responses are condensed
  for display.}
  \label{fig:robot}
\end{figure*}

\subsection{Embodied Question Answering}
\label{sec:eqa}

Embodied question answering (EQA) requires an agent to explore an environment,
gather relevant visual evidence, and answer a question using as little motion
as possible. We evaluate \asena on HM-EQA and MT-HM3D using their released
initial poses~\cite{ren2024exploreeqa,zhai2025memoryeqa}. Following the
Explore-EQA teleportation protocol, each episode has a nominal step budget of
$3\sqrt{A}$ for a scene of area $A$. We report answer accuracy and normalized
exploration steps, defined as the fraction of this nominal budget consumed by
the agent.

\paragraph{Hand-designed reference configurations}
Our EQA agent composes navigation, visual inspection, object search, place
memory, and answer-generation tools. We first construct two hand-designed
configurations as reference points. The \emph{quality} configuration
prioritizes collecting sufficient evidence to answer correctly and receives
no explicit instruction to minimize exploration. The \emph{efficient}
configuration is prompted to treat exploration steps as a limited resource
and gather evidence more selectively. On HM-EQA, we give the efficient agent
a target budget equal to 30\% of the nominal allowance, while retaining the
same simulator hard limit for both configurations. These configurations
provide manually designed operating points for accuracy and efficiency rather
than a controlled ablation of a single component.

\paragraph{Workspace-evolution protocol}
We next study whether the agent can improve this tradeoff by revising its
workspace. We construct fixed evolution pools containing 398 HM-EQA questions
and 1,183 MT-HM3D questions from training scenes. During each pass, sixteen
workers sample 50 questions from each benchmark, execute them using private
workspaces, and reflect on their outcomes. A separate improver aggregates
their traces and edits to update the canonical workspace, while all model
weights remain fixed. Questions not sampled by the workers are used for
monitoring, and their aggregate results are also provided to the improver.

\paragraph{Workspace evolution improves the tradeoff between cost and accuracy}
With workspace frozen, pass 17 achieves the highest HM-EQA accuracy
(81.2\%), while pass 4 performs best on MT-HM3D (66.5\%)
(\tabref{tab:eqaevolve}). At low reasoning effort, evolved workspaces use
steps comparable to the hand-designed efficient configuration while achieving
higher accuracy (\tabref{tab:eqa}). Improvements vary across revisions as the
workspace adds new routines and refines existing guidance. These revisions
produce different tradeoffs between accuracy and exploration efficiency.

\paragraph{Learned navigation reduces exploration}
In a paired ablation on 132 MT-HM3D questions, \capravln reduces median
normalized exploration steps by approximately 19\% with little change in
accuracy. Consistent with our previous R2R and RxR results, learned navigation improves
efficiency by reducing repeated agent interaction.

\subsection{Real-World Applications}
\label{sec:robot}

We present four demonstrations of \asena on a Unitree G1
(\figref{fig:robot}). All four operate without pre-built maps and combine
navigation with search, user interaction, or object delivery. The agent gathers
visual and geometric evidence and maintains spatial memory. It writes or revises
programs that connect perception and planning with physical actions
(\figref{fig:teaser}).

The three search missions (\figref{fig:robot}A--C) take place in previously unseen
environments. Search targets are initially outside the robot's view. In the
restroom mission, the robot locates doors near the elevator lobby. It uses its
online map to identify a shorter route from the user and returns with directions.
In the pantry mission, it locates the room, inspects a dispenser
and brewer, and reports two coffee machines. In the snack-comparison mission, it
inspects a reference snack and finds a vending machine. It checks whether the
same item is available, then returns to answer the user with a generated
whole-body gesture. These tasks combine online search and spatial reasoning with evidence
collection, memory, and multimodal communication.

The fetch-and-handoff task (\figref{fig:robot}D) demonstrates humanoid mobile
manipulation by combining navigation, object pickup, carrying while walking,
and handoff. After reaching the pantry, the agent uses a
head-camera observation to locate the can and ground a grasp. It inspects an
arm-motion plan, then uses a YAM gripper to pick up the can. The robot
walks back while carrying it. A separately programmed handoff uses inverse kinematics
to position it above the recipient's hand before releasing the gripper. This
qualitative demonstration combines navigation, grasp planning, and arm control
through the same programming interface.

%% file: tables/icra-nv/eqa.tex
\begin{table}[t]
\centering
\caption{Embodied question answering.
Accuracy (\%) and mean normalized steps with differing coverage.}
\label{tab:eqa}
\footnotesize
\setlength{\tabcolsep}{2pt}
\begin{tabularx}{\columnwidth}{l*{4}{Y}}
\toprule
 & \multicolumn{2}{c}{HM-EQA} & \multicolumn{2}{c}{MT-HM3D} \\
\cmidrule(lr){2-3}\cmidrule(lr){4-5}
Method & Acc.$\uparrow$ & Steps$\downarrow$ & Acc.$\uparrow$ & Steps$\downarrow$ \\
\midrule
Explore-EQA~\cite{ren2024exploreeqa} & 58.4 & .52 & 35.1 & .64 \\
3D-Mem~\cite{yang2025threedmem} & 50.4 & .63 & -- & -- \\
Fine-EQA~\cite{jiang2025fineeqa} & 56.0 & .54 & -- & -- \\
GraphEQA~\cite{saxena2025grapheqa} & 63.5 & .20 & 45.6 & .45 \\
MemoryEQA (Qwen2VL-7B)~\cite{zhai2025memoryeqa} & -- & -- & 51.2 & .40 \\
MemoryEQA (GPT-4o)~\cite{zhai2025memoryeqa} & 63.4 & .40 & 55.1 & .41 \\
FAST-EQA~\cite{zhang2026fasteqa} & 69.2 & .65 & 50.5 & .52 \\
Qwen-RobotNav~\cite{qwen2026robotnav} & 76.7 & .15 & 54.4 & .19 \\
\midrule
\rowcolor{asenashade}
\asena base (low effort) & 70.9 & .15 & 64.8 & \underline{.10} \\
\rowcolor{asenashade}
\asena hand-designed (efficient) & 72.2 & \textbf{.11} & 63.8 & .11 \\
\rowcolor{asenashade}
\asena hand-designed (quality) & \underline{78.8} & .22 & \textbf{67.1} & .24 \\
\rowcolor{asenashade}
\asena self-evolved 17 (low) & 74.8 & \underline{.13} & 64.9 & \textbf{.08} \\
\rowcolor{asenashade}
\asena self-evolved 17 (high) & \textbf{81.2} & .25 & \underline{65.5} & .22 \\
\bottomrule
\end{tabularx}

\end{table}

%% file: tables/icra-nv/eqa-workspace.tex
\begin{table}[t]
  \centering
  \caption{Frozen EQA workspaces: accuracy (\%) and normalized exploration steps. Shading marks the final workspace.}
  \label{tab:eqaevolve}
  \footnotesize
  \setlength{\tabcolsep}{3.2pt}
  \begin{tabularx}{\columnwidth}{ll*{4}{Y}}
    \toprule
    & & \multicolumn{2}{c}{HM-EQA} & \multicolumn{2}{c}{MT-HM3D} \\
    \cmidrule(lr){3-4}\cmidrule(lr){5-6}
    Pass & Effort & Acc.$\uparrow$ & Steps$\downarrow$ & Acc.$\uparrow$ & Steps$\downarrow$ \\
    \midrule
    Base & Low & 70.9 & .153 & 64.8 & .101 \\
    4 & Low & 73.4 & \underline{.136} & \textbf{66.5} & \underline{.091} \\
    9 & Low & \underline{75.8} & .147 & \underline{65.5} & .116 \\
    11 & Low & 75.4 & .152 & 65.3 & .114 \\
    \rowcolor{asenashade}
    17 & Low & 74.8 & \textbf{.127} & 64.9 & \textbf{.076} \\
    \rowcolor{asenashade}
    17 & High & \textbf{81.2} & .253 & \underline{65.5} & .224 \\
    \bottomrule
  \end{tabularx}
\end{table}

%% file: section/icra-nv/05-discussion.tex
\section{Discussion and Conclusion}
\asena unifies coding-agent programming, robot execution, and persistent
skills. It improves recurring-task performance while model weights remain fixed. Our
optional 4B monocular \capravln policy, trained on curated atomic-navigation
data, achieves leading R2R/RxR success rates with fewer agent interactions.
\asena also achieves leading EQA accuracy. On G1, online programming extends
navigation to search, inspection, and interaction. The agent synthesizes
whole-body responses online.

Several limitations remain. The G1 demonstrations
provide qualitative evidence of capability composition rather than a
systematic measure of reliability across objects, layouts, and users.
Real-world tasks take 10--20 minutes, making faster reasoning and execution
important for practical use. Future work should evaluate retained skills on
new tasks, measure navigation and manipulation reliability over repeated
real-world trials, and strengthen safeguards for dynamic environments.
Another direction is to extend workspace evolution to both learned policy and
agent updates.

\section*{Acknowledgment}
We thank Jarred Travers, Amanpreet Singh, Peter Pham, Jinhyung (David) Park, Xiangchen Tian, and Tingwu Wang for their help with real robot infrastructure.

%% file: section/07-supplement-nv.tex
\section{Supplementary Material}

\subsection{Atomic Instruction Format and Training Data}
\label{sec:supp:atomic_training}

\input{tables/atomic_format-nv}
\input{tables/training_data-nv}

\paragraph{Training Sampling}
\label{sec:supp:training_scope}
Training uses 64 devices and an effective batch size of 512. Each device processes
three trajectory examples and one VQA example per step. We accumulate gradients
over two steps. The reported checkpoint uses fewer than 10K optimizer steps
and fewer than 5M training samples, including repeats.
To emphasize stopping behavior, we add five extra copies of each eligible
trajectory-ending window. The pools also include alternative views, CoT variants,
and repeated observations.

\subsection{Jetson Thor Quantization and Runtime}
\label{sec:supp:quantization}

To run the policy on the G1's Jetson AGX Thor, we apply post-training FP8
quantization with NVIDIA ModelOpt. We build inference engines with
TensorRT-Edge-LLM. These measurements use an earlier \capravln\ checkpoint at
step 8,400, separate from the benchmark policy in \Cref{tab:vln-policy}.
Every FP8 variant quantizes the language model. Variants labeled ``BF16 vision''
keep the 415M-parameter vision encoder in BF16. We leave the embedding and output
projection unquantized in all variants.

Navigation-only engines need to predict motion tokens rather than general text.
We therefore reduce the output vocabulary from 152,596 to 768 entries while
retaining all 660 trajectory tokens. This removes 388.7M parameters from the
output projection evaluated at each decoding step.

\paragraph{Measurement protocol}
All variants use batch size 1, the same eight-frame mixed-resolution history,
and 32 greedy decoding steps. BF16 values are warm medians of five runs.
Prefill is time to first token minus visual encoding. Decode scales the
remaining 31 tokens to 32 steps. TensorRT-Edge-LLM visual and prefill values
are 10-run means. Decode is the median of five runs.

Measured BF16 end-to-end latency is 2,735\,ms. The BF16 model occupies
8.28\,GiB of allocated memory after loading. This is not a peak runtime
measurement.

\input{tables/quantization-nv}

\Cref{tab:quant_latency} reports latency for each inference component. FP8 with the
reduced vocabulary lowers the sum of component latencies from 2,806 to 877\,ms,
a 3.20$\times$ speedup over PyTorch BF16 under the same input settings.
Decoding remains the largest cost at 683.4\,ms, or 78\% of the FP8 total.
Among the BF16-vision variants, reducing the vocabulary cuts decoding latency
from 847.4 to 694.1\,ms (18.1\%). Visual encoding and prefill latencies stay
nearly unchanged.

\Cref{tab:quant_fidelity} compares quantized and BF16 predictions on held-out
inputs. Keeping the vision encoder in BF16 improves
agreement for both the first waypoint and the full trajectory. These checks
measure output consistency. Navigation success requires separate rollout
evaluation. On-device checks also confirmed valid navigation-token IDs and
correctly formatted eight-waypoint trajectories.

\clearpage
\input{section/atomic-gallery-nv}
\clearpage
\input{section/atomic-examples-nv}
\clearpage
\input{section/08-hardware-nv}

%% file: tables/atomic_format-nv.tex
\begin{table}[H]
    \centering
    \footnotesize
    \setlength{\tabcolsep}{6pt}
    \begin{tabularx}{\linewidth}{@{}ll>{\raggedright\arraybackslash}X@{}}
        \toprule
        \textbf{Stored field} & \textbf{Type} & \textbf{Meaning} \\
        \midrule
        \makecell[l]{\textbf{goal\_type}\\\textbf{goal\_value}} & category; text &
            Goal category and text. Recaptioned records can store just the destination
            phrase, while raw instructions and generated atoms retain the complete
            instruction. Categories include object, area, ego, region and point. \\
        \textbf{route\_steps} & ordered list &
            Ordered directions and landmark cues, when separately annotated. \\
        \textbf{constraints} & list &
            Avoidance, traversal, or relative-position requirements. \\
        \textbf{end\_pose} & anchor list &
            Target, spatial relation, and distance derived from scene geometry,
            relative to an object, a room, or the robot's starting position. \\
        \textbf{quality} & 1--5 &
            Quality annotation or assigned setting used to condition the policy.
            R2R/RxR evaluation uses 5. \\
        \bottomrule
    \end{tabularx}
    \caption{Stored navigation annotations. The policy receives goal text, quality,
    and optional end-pose anchors (\Cref{tab:atomic_examples_nv}).
    Coordinate-based point goals use a separate input format.}
    \label{tab:atomic_format}
\end{table}

%% file: tables/training_data-nv.tex
\begin{table}[H]
    \centering
    \footnotesize
    \setlength{\tabcolsep}{6pt}
    \begin{tabular}{llcc}
        \toprule
        \textbf{Dataset} & \textbf{Supervision} & \textbf{Weighted pool (M)} & \textbf{Environments} \\
        \midrule
        R2R~\citep{anderson2018r2r}             & traj, pixel, CoT & 1.502 & MP3D \\
        R2R (augmented)                         & traj, pixel      & 3.290 & MP3D \\
        FGR2R~\citep{hong2020subinstruction}    & traj, pixel      & 0.704 & MP3D \\
        RxR~\citep{ku2020rxr}                   & traj, pixel, CoT & 4.191 & MP3D \\
        RxR (augmented)                         & traj, pixel      & 4.836 & MP3D \\
        ScaleVLN~\citep{wang2023scalevln}       & traj, pixel, CoT & 3.120 & HM3D \\
        SRDF~\citep{wang2025srdf}               & traj, pixel      & 3.916 & HM3D \\
        ObjectNav MP3D v3                       & traj, pixel      & 0.707 & MP3D \\
        ObjectNav HM3D v3~\citep{ramakrishnan2021hm3d} & traj, pixel & 2.888 & HM3D \\
        Nav-atoms (synthetic, MP3D)             & traj, pixel      & 1.443 & MP3D \\
        Nav-atoms (synthetic, HM3D)             & traj, pixel      & 1.825 & HM3D \\
        VLN-MME + other general VQA             & VQA              & 0.964 & various \\
        SenseNova-SI~\citep{sensenova-si}       & VQA              & 8.164 & various \\
        \midrule
        \textbf{Trajectory pool}               &                  & \textbf{28.422} & \\
        \textbf{VQA pool}                      &                  & \textbf{9.128} & \\
        \bottomrule
    \end{tabular}
    \caption{Training pools for \capravln. Pool sizes account for quality and
    backward-action filtering, alternative views, chain-of-thought (CoT) variants,
    repeated trajectory-ending windows, and sampling weights. Counts are rounded
    independently and include repeated examples. They describe the available pools,
    not the number of samples consumed during training.
    Sampling details appear in \Cref{sec:supp:training_scope}.}
    \label{tab:training_data}
\end{table}

%% file: tables/quantization-nv.tex
\begin{table}[H]
    \centering
    \footnotesize
    \setlength{\tabcolsep}{4pt}
    \begin{tabular}{lrrrrrr}
        \toprule
        \textbf{Variant} & \textbf{Export size} & \textbf{Visual} & \textbf{Prefill} &
        \textbf{Decode (32)} & \textbf{Total} & \textbf{Speedup} \\
        & \textbf{(GiB)} & \multicolumn{4}{c}{\textbf{Latency (ms)}} & \\
        \midrule
        PyTorch BF16                 & 9.01 & 167.3 & 363.4 & 2,275.5 & 2,806 & 1.00$\times$ \\
        FP8, BF16 vision              & 5.63 & 73.3 & 128.7 & 847.4 & 1,050 & 2.67$\times$ \\
        FP8, BF16 vision, reduced vocab & 4.91 & 78.3 & 125.5 & 694.1 & 898 & 3.13$\times$ \\
        FP8, reduced vocab            & 4.53 & 64.9 & 129.1 & 683.4 & 877 & 3.20$\times$ \\
        \bottomrule
    \end{tabular}
    \caption{Policy inference on the Unitree G1's Jetson AGX Thor. These measurements
    use an earlier checkpoint, separate from the benchmark policy. Export size
    describes the corresponding ONNX files. Totals sum the three latency
    components. Speedups are computed from unrounded sums.}
    \label{tab:quant_latency}
\end{table}

\begin{table}[H]
    \centering
    \footnotesize
    \setlength{\tabcolsep}{8pt}
    \begin{tabular}{lcc}
        \toprule
        \textbf{Variant} & \textbf{First-waypoint agreement} &
        \textbf{Exact-trajectory agreement} \\
        \midrule
        FP8                       & 90.23\% & 75.39\% \\
        FP8, BF16 vision          & 91.41\% & 82.81\% \\
        \bottomrule
    \end{tabular}
    \caption{Prediction agreement between the earlier quantized model and its BF16
    version on 256 held-out inputs. First-waypoint agreement requires the first
    predicted $(x,y,\theta)$ tuple to match. Exact-trajectory agreement requires the
    entire predicted trajectory to match. These metrics measure output consistency,
    not navigation success.}
    \label{tab:quant_fidelity}
\end{table}

%% file: section/atomic-gallery-nv.tex
\subsection{Atomic Navigation Dataset Samples}
\label{sec:supp:atomic_gallery}
\begingroup
\setlength{\intextsep}{4pt}
\begin{figure}[H]
  \centering
  \input{figures/icra-nv/atomic-training-gallery}
  \caption{\textbf{Atomic navigation training examples.} Each observation is paired
  with an instruction and an overlaid trajectory. The examples span metric
  motion, rotation, relative goals, alignment, room goals, doors, corridors, and
  floor changes.}
  \label{fig:atomic-training-gallery}
\end{figure}
\endgroup

%% file: figures/icra-nv/atomic-training-gallery.tex
\begingroup
\input{figures/icra-nv/atomic-samples-style}
\resizebox{\linewidth}{!}{%
\begin{tikzpicture}[x=1bp,y=-1bp]
\path[use as bounding box] (0,0) rectangle (470,525);
\begin{scope}[shift={(0,0)}]
\node[anchor=north west,inner sep=0,text=atomicGreen,font=\atomicBold{8.5}]
  at (0,0) {Metric motion};
\begin{scope}
\clip[rounded corners=2bp] (0,12) rectangle (150,96.375);
\node[anchor=north west,inner sep=0] at (0,12)
  {\includegraphics[width=150bp]{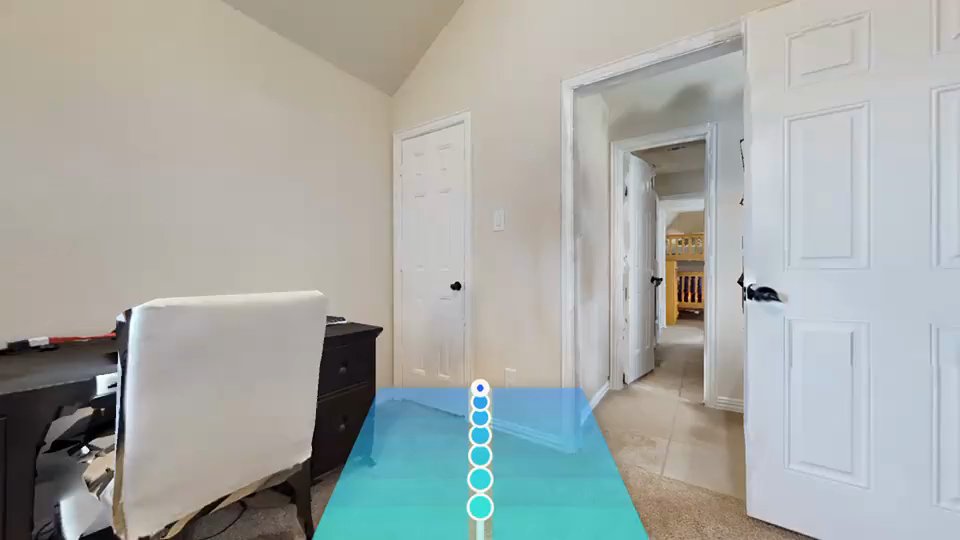}};
\end{scope}
\node[anchor=north west,inner sep=0,text=atomicInk,text width=150bp,
  align=left,font=\atomicRegular{8.5}]
  at (0,100.375) {Stop after you take 4 steps forward.};
\end{scope}
\begin{scope}[shift={(160,0)}]
\node[anchor=north west,inner sep=0,text=atomicGreen,font=\atomicBold{8.5}]
  at (0,0) {Pure rotation};
\begin{scope}
\clip[rounded corners=2bp] (0,12) rectangle (150,96.375);
\node[anchor=north west,inner sep=0] at (0,12)
  {\includegraphics[width=150bp]{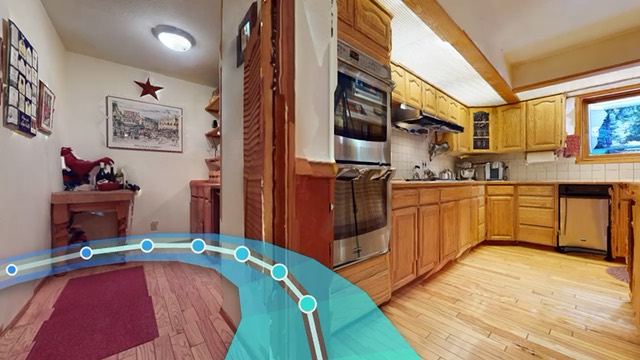}};
\end{scope}
\node[anchor=north west,inner sep=0,text=atomicInk,text width=150bp,
  align=left,font=\atomicRegular{8.5}]
  at (0,100.375) {You will need to turn all the way around to face behind you.};
\end{scope}
\begin{scope}[shift={(320,0)}]
\node[anchor=north west,inner sep=0,text=atomicGreen,font=\atomicBold{8.5}]
  at (0,0) {Heading-relative goal};
\begin{scope}
\clip[rounded corners=2bp] (0,12) rectangle (150,96.375);
\node[anchor=north west,inner sep=0] at (0,12)
  {\includegraphics[width=150bp]{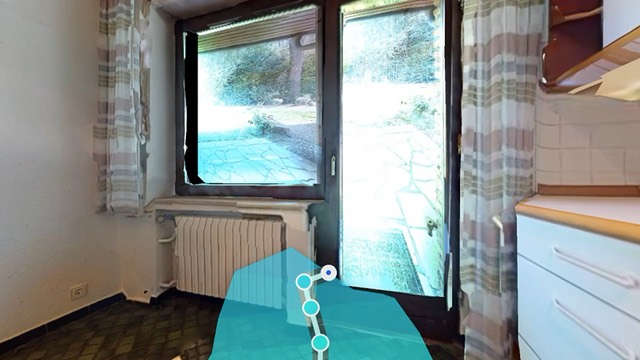}};
\end{scope}
\node[anchor=north west,inner sep=0,text=atomicInk,text width=150bp,
  align=left,font=\atomicRegular{8.5}]
  at (0,100.375) {Walk over to the built-in dishwasher off to your left.};
\end{scope}
\begin{scope}[shift={(0,135)}]
\node[anchor=north west,inner sep=0,text=atomicGreen,font=\atomicBold{8.5}]
  at (0,0) {Fine alignment};
\begin{scope}
\clip[rounded corners=2bp] (0,12) rectangle (150,96.375);
\node[anchor=north west,inner sep=0] at (0,12)
  {\includegraphics[width=150bp]{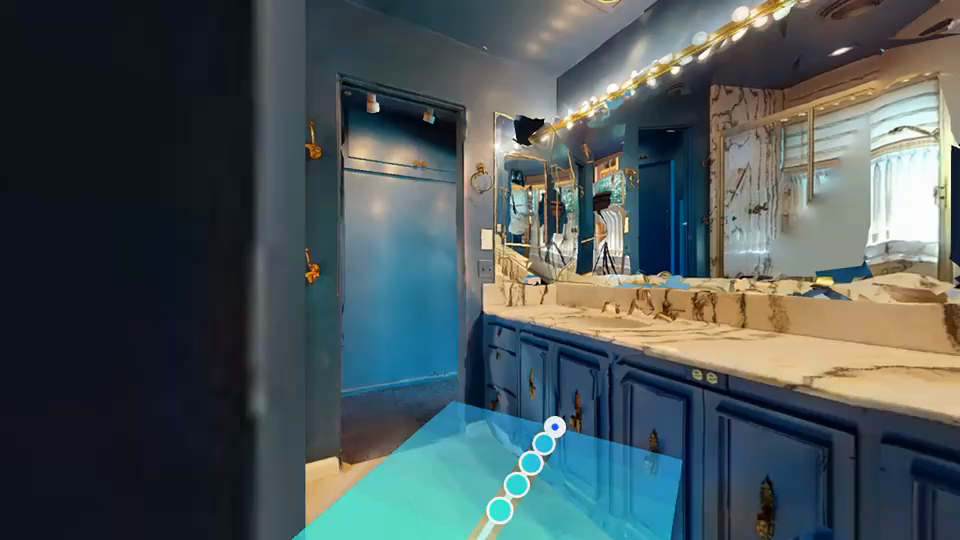}};
\end{scope}
\node[anchor=north west,inner sep=0,text=atomicInk,text width=150bp,
  align=left,font=\atomicRegular{8.5}]
  at (0,100.375) {Sidestep right toward the bathroom exhaust fan.};
\end{scope}
\begin{scope}[shift={(160,135)}]
\node[anchor=north west,inner sep=0,text=atomicGreen,font=\atomicBold{8.5}]
  at (0,0) {Room goal};
\begin{scope}
\clip[rounded corners=2bp] (0,12) rectangle (150,96.375);
\node[anchor=north west,inner sep=0] at (0,12)
  {\includegraphics[width=150bp]{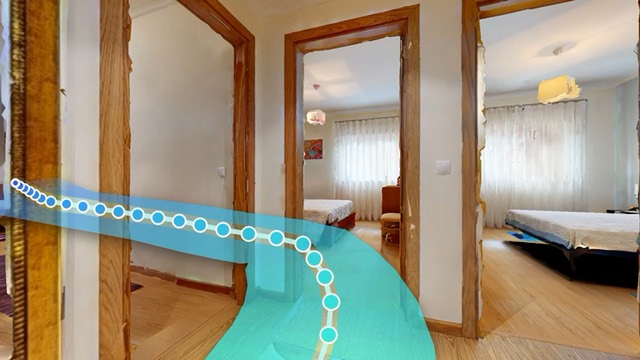}};
\end{scope}
\node[anchor=north west,inner sep=0,text=atomicInk,text width=150bp,
  align=left,font=\atomicRegular{8.5}]
  at (0,100.375) {Stop there once you find a bedroom.};
\end{scope}
\begin{scope}[shift={(320,135)}]
\node[anchor=north west,inner sep=0,text=atomicGreen,font=\atomicBold{8.5}]
  at (0,0) {Door traversal};
\begin{scope}
\clip[rounded corners=2bp] (0,12) rectangle (150,96.375);
\node[anchor=north west,inner sep=0] at (0,12)
  {\includegraphics[width=150bp]{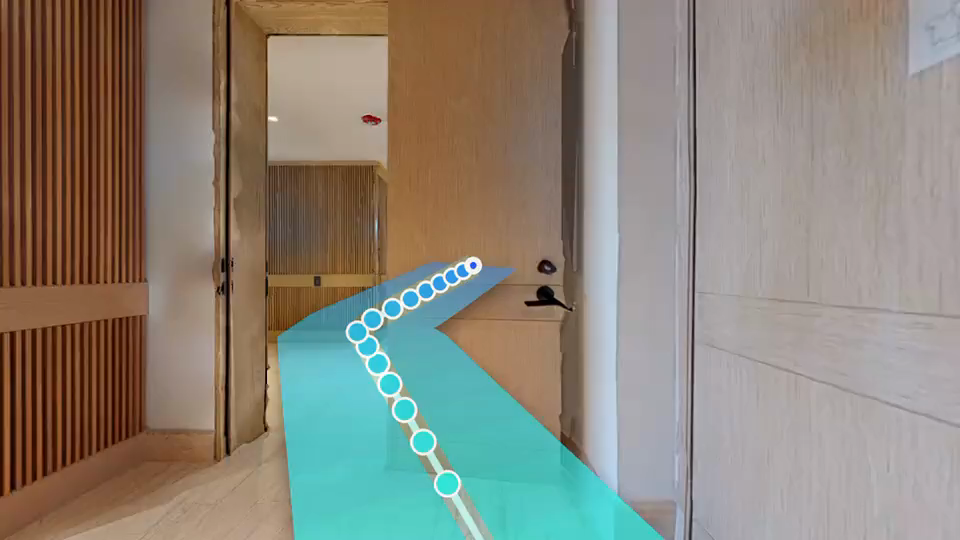}};
\end{scope}
\node[anchor=north west,inner sep=0,text=atomicInk,text width=150bp,
  align=left,font=\atomicRegular{8.5}]
  at (0,100.375) {Step into the kitchen through the door.};
\end{scope}
\begin{scope}[shift={(0,270)}]
\node[anchor=north west,inner sep=0,text=atomicGreen,font=\atomicBold{8.5}]
  at (0,0) {Corridor following};
\begin{scope}
\clip[rounded corners=2bp] (0,12) rectangle (150,96.375);
\node[anchor=north west,inner sep=0] at (0,12)
  {\includegraphics[width=150bp]{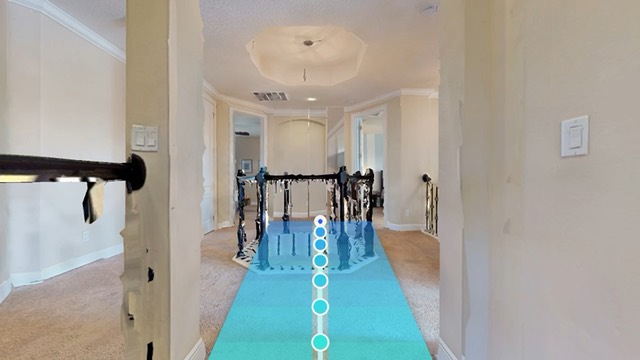}};
\end{scope}
\node[anchor=north west,inner sep=0,text=atomicInk,text width=150bp,
  align=left,font=\atomicRegular{8.5}]
  at (0,100.375) {Head down the corridor and stop at the end.};
\end{scope}
\begin{scope}[shift={(160,270)}]
\node[anchor=north west,inner sep=0,text=atomicGreen,font=\atomicBold{8.5}]
  at (0,0) {Floor change};
\begin{scope}
\clip[rounded corners=2bp] (0,12) rectangle (150,96.375);
\node[anchor=north west,inner sep=0] at (0,12)
  {\includegraphics[width=150bp]{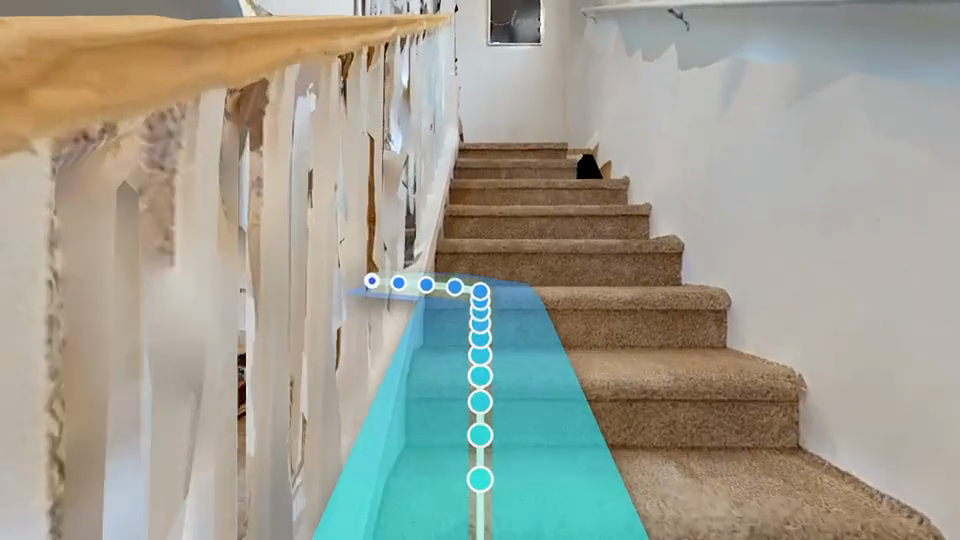}};
\end{scope}
\node[anchor=north west,inner sep=0,text=atomicInk,text width=150bp,
  align=left,font=\atomicRegular{8.5}]
  at (0,100.375) {Head upstairs.};
\end{scope}
\begin{scope}[shift={(320,270)}]
\node[anchor=north west,inner sep=0,text=atomicGreen,font=\atomicBold{8.5}]
  at (0,0) {Heading-relative goal};
\begin{scope}
\clip[rounded corners=2bp] (0,12) rectangle (150,96.375);
\node[anchor=north west,inner sep=0] at (0,12)
  {\includegraphics[width=150bp]{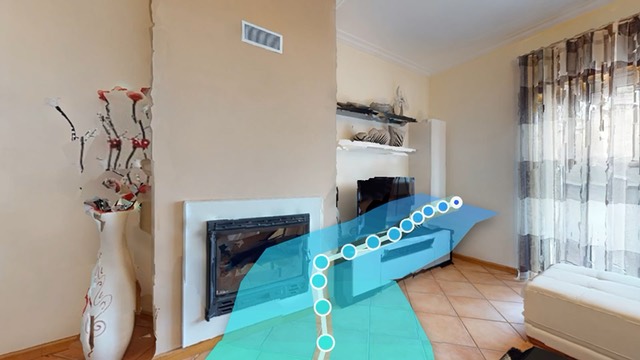}};
\end{scope}
\node[anchor=north west,inner sep=0,text=atomicInk,text width=150bp,
  align=left,font=\atomicRegular{8.5}]
  at (0,100.375) {Turn around and walk to the white wall-mounted cabinet right behind you.};
\end{scope}
\begin{scope}[shift={(0,405)}]
\node[anchor=north west,inner sep=0,text=atomicGreen,font=\atomicBold{8.5}]
  at (0,0) {Metric motion};
\begin{scope}
\clip[rounded corners=2bp] (0,12) rectangle (150,96.375);
\node[anchor=north west,inner sep=0] at (0,12)
  {\includegraphics[width=150bp]{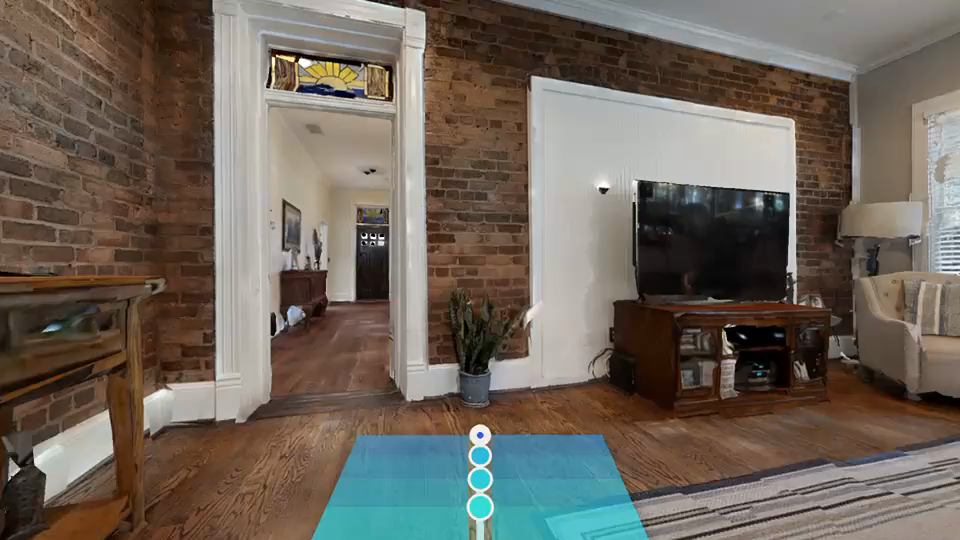}};
\end{scope}
\node[anchor=north west,inner sep=0,text=atomicInk,text width=150bp,
  align=left,font=\atomicRegular{8.5}]
  at (0,100.375) {Can you move forward 0.5 meters?};
\end{scope}
\begin{scope}[shift={(160,405)}]
\node[anchor=north west,inner sep=0,text=atomicGreen,font=\atomicBold{8.5}]
  at (0,0) {Door traversal};
\begin{scope}
\clip[rounded corners=2bp] (0,12) rectangle (150,96.375);
\node[anchor=north west,inner sep=0] at (0,12)
  {\includegraphics[width=150bp]{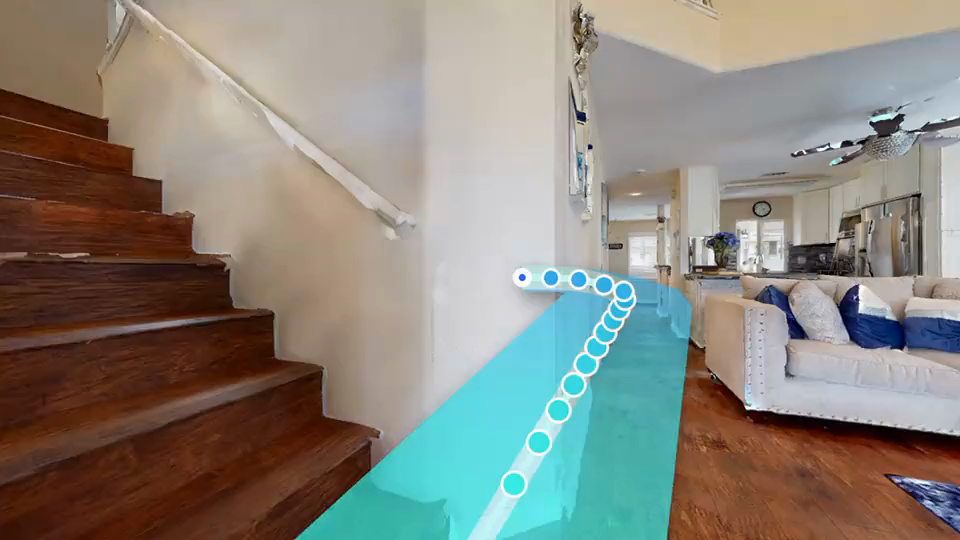}};
\end{scope}
\node[anchor=north west,inner sep=0,text=atomicInk,text width=150bp,
  align=left,font=\atomicRegular{8.5}]
  at (0,100.375) {Step into the hallway.};
\end{scope}
\begin{scope}[shift={(320,405)}]
\node[anchor=north west,inner sep=0,text=atomicGreen,font=\atomicBold{8.5}]
  at (0,0) {Floor change};
\begin{scope}
\clip[rounded corners=2bp] (0,12) rectangle (150,96.375);
\node[anchor=north west,inner sep=0] at (0,12)
  {\includegraphics[width=150bp]{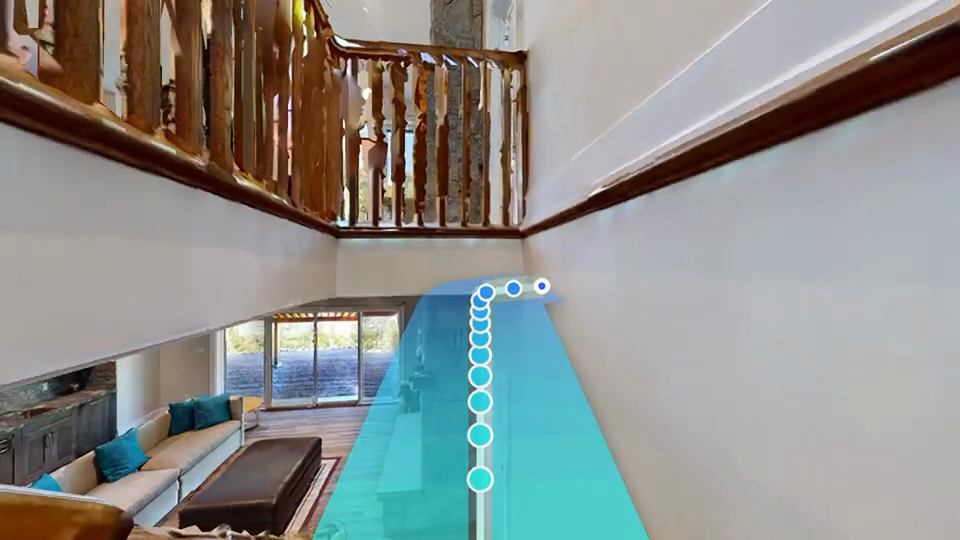}};
\end{scope}
\node[anchor=north west,inner sep=0,text=atomicInk,text width=150bp,
  align=left,font=\atomicRegular{8.5}]
  at (0,100.375) {You should head down the stairs.};
\end{scope}
\end{tikzpicture}%
}
\endgroup

%% file: section/atomic-examples-nv.tex
\subsection{Navigation Instructions and Policy Inputs}
\label{sec:supp:atomic_examples}
\begingroup
\setlength{\parskip}{3pt}
\setlength{\intextsep}{6pt}
\captionsetup{font=small}

\Cref{tab:atomic_examples_nv} shows the navigation instruction format that the
policy receives. Each input contains the goal text and a quality setting.
An optional end-pose anchor describes where the robot should finish relative to
an object, a room, or its starting position.

\input{tables/atomic_examples-nv}

Directions, route cues, and constraints stay in the instruction text.
We express end-pose distances in meters and round them to two decimal places.
A room goal uses the relation \texttt{inside} with distance 0.0.
All five examples use quality 5. The policy receives this instruction input
together with its visual history.
\endgroup

%% file: tables/atomic_examples-nv.tex
\begin{table}[H]
\centering
\small
\setlength{\tabcolsep}{7pt}
\caption{\textbf{Navigation instructions and policy inputs.} Each JSON input
contains the goal text and a quality setting, with an optional end-pose anchor
that specifies where the robot should finish.}
\label{tab:atomic_examples_nv}
\begin{tabularx}{\linewidth}{@{}p{0.44\linewidth}X@{}}
\toprule
Navigation instruction & Policy input (JSON) \\
\midrule
\textbf{Route following}\par Descend down the stairs. Turn left and stop in the room.  & \begin{minipage}[t]{\linewidth}\raggedright\ttfamily\fontsize{8}{9.5}\selectfont \{"goal": "Descend down the stairs. Turn left and stop in the room. ",\\ "quality": 5\}\end{minipage} \\
\addlinespace[8pt]
\textbf{Metric motion}\par Could you move forward 1 meters for me? & \begin{minipage}[t]{\linewidth}\raggedright\ttfamily\fontsize{8}{9.5}\selectfont \{"goal": "Could you move forward 1 meters for me?",\\ "quality": 5,\\ "end\_pose": [\{\\   "target": "agent\_start",\\   "relation": "forward",\\   "distance\_m": 1.0\\ \}]\}\end{minipage} \\
\addlinespace[8pt]
\textbf{Relative object goal}\par Walk over to the built-in dishwasher off to your left. & \begin{minipage}[t]{\linewidth}\raggedright\ttfamily\fontsize{8}{9.5}\selectfont \{"goal": "Walk over to the built-in dishwasher off to your left.",\\ "quality": 5,\\ "end\_pose": [\{\\   "target": "built-in dishwasher",\\   "relation": "near",\\   "distance\_m": 0.55\\ \}]\}\end{minipage} \\
\addlinespace[8pt]
\textbf{Room goal}\par Stop there once you find a bedroom. & \begin{minipage}[t]{\linewidth}\raggedright\ttfamily\fontsize{8}{9.5}\selectfont \{"goal": "Stop there once you find a bedroom.",\\ "quality": 5,\\ "end\_pose": [\{\\   "target": "bedroom",\\   "relation": "inside",\\   "distance\_m": 0.0\\ \}]\}\end{minipage} \\
\addlinespace[8pt]
\textbf{Floor change}\par Make your way downstairs to the hallway. & \begin{minipage}[t]{\linewidth}\raggedright\ttfamily\fontsize{8}{9.5}\selectfont \{"goal": "Make your way downstairs to the hallway.",\\ "quality": 5,\\ "end\_pose": [\{\\   "target": "hallway",\\   "relation": "inside",\\   "distance\_m": 0.0\\ \}]\}\end{minipage} \\
\bottomrule\end{tabularx}\end{table}

%% file: section/08-hardware-nv.tex
\subsection{G1 Hardware and Navigation Settings}
\label{sec:supp:hardware}
\begingroup
\setlength{\intextsep}{4pt}
\setlength{\parskip}{4pt}
\captionsetup{font=small,skip=4pt}

The G1 carries a mast-mounted panoramic camera and an inverted Livox Mid-360
LiDAR (\Cref{fig:g1_hardware_nv}). A Jetson AGX Thor runs the local sensor and
motion services. The camera interface supports both the Insta360 X5 and
Ricoh Theta X. Camera yaw correction sets the panorama's horizontal orientation.
The robot's head and body partially occlude the inverted LiDAR's view.
For manipulation, the G1 uses a calibrated head OAK RGB/stereo camera and optional
YAM grippers with fin-ray fingers. Gripper control runs separately from SONIC
body motion. \Cref{tab:g1_hardware_nv} lists the sensor and state outputs.

\input{figures/g1-hardware-nv}

\input{tables/g1-hardware-nv}

\Cref{tab:g1_drive_nv} summarizes the navigation settings and command limits.
The slow-walk settings control motion speed, stopping tolerances, command size,
and timeouts. They can be adjusted at runtime. A separate walking gait is also
available.

\input{tables/g1-drive-nv}
\endgroup

%% file: figures/g1-hardware-nv.tex
\begin{figure}[H]
\centering
\scalebox{0.9}{%
\begin{tikzpicture}[x=1pt,y=1pt,
    every node/.style={font=\small,inner sep=0pt},
    callout/.style={draw=srLinkGreen,line width=0.7pt}]
  \path[use as bounding box] (0,0) rectangle (452,176);
  \node[anchor=west,font=\small\bfseries] at (0,170) {A\quad Front view};
  \node[anchor=west,font=\small\bfseries] at (236,170) {B\quad Side view};
  \begin{scope}
    \clip (0,0) rectangle (107.744,155);
    \node[anchor=south west] at (-109.634,-54.817)
      {\includegraphics[width=483.902pt]{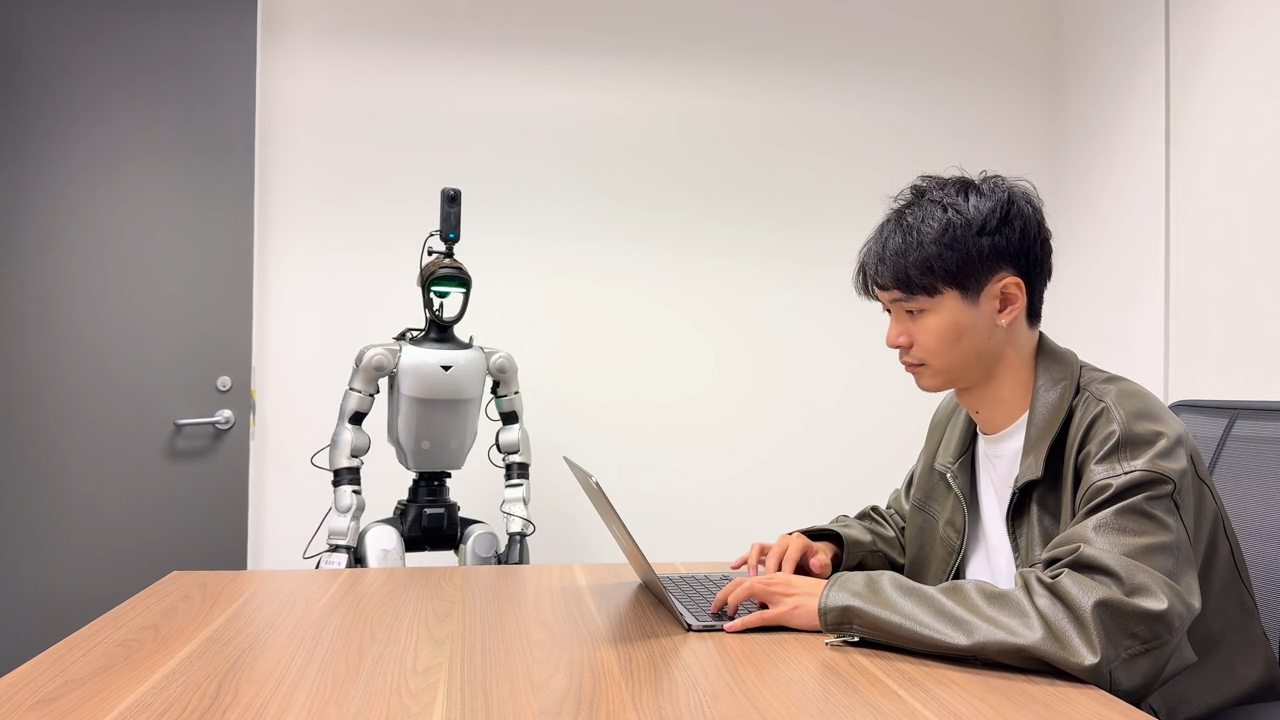}};
  \end{scope}
  \begin{scope}
    \clip (236,0) rectangle (339.941,155);
    \node[anchor=south west] at (155.765,-49.235)
      {\includegraphics[width=466.824pt]{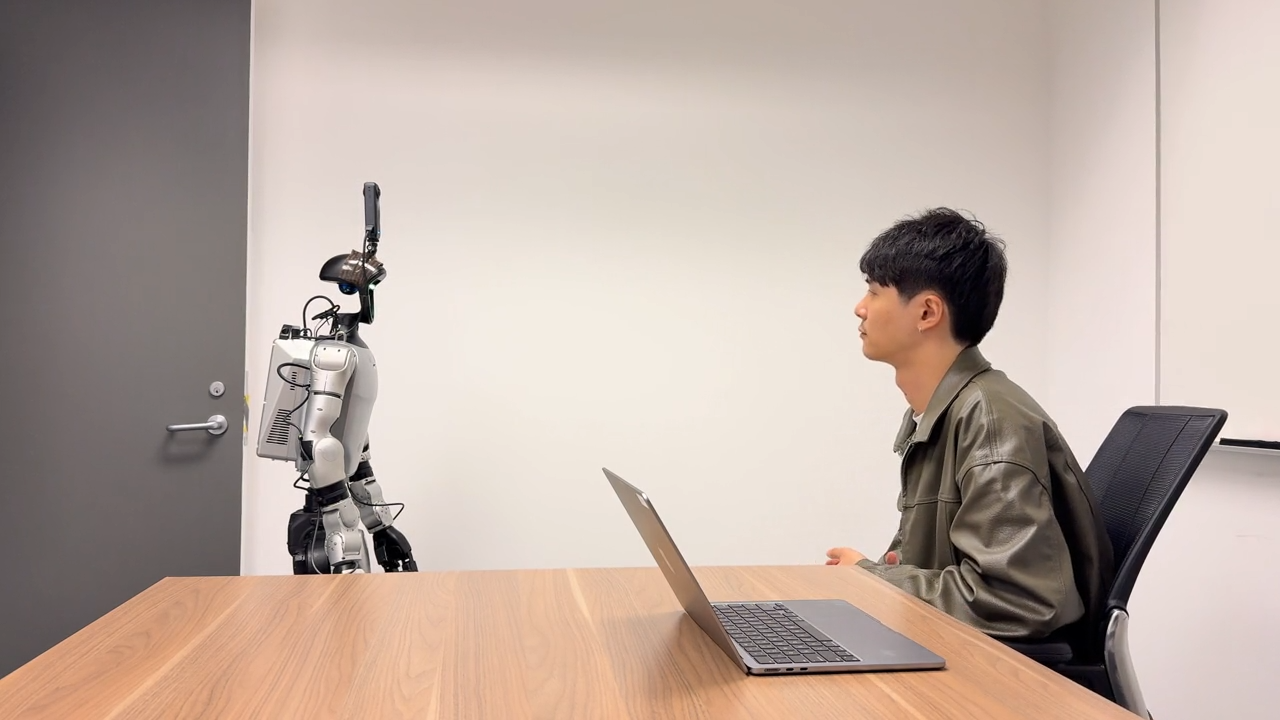}};
  \end{scope}
  \draw[callout] (62,139) -- (101,139) -- (112,143);
  \fill[srLinkGreen] (62,139) circle (1.8pt);
  \node[anchor=west,align=left] at (116,143) {360\degree\ camera\\on a mast};
  \draw[callout] (61,106) -- (99,106) -- (112,100);
  \fill[srLinkGreen] (61,106) circle (1.8pt);
  \node[anchor=west,align=left] at (116,100) {Lidar inside\\the head shell};
  \draw[callout] (264,69) -- (335,69) -- (348,75);
  \fill[srLinkGreen] (264,69) circle (1.8pt);
  \node[anchor=west,align=left] at (352,75) {Rear\\enclosure};
\end{tikzpicture}%
}
\caption{\textbf{G1 sensor mounting.} Earlier configuration, before the head
RGB/stereo camera and optional YAM grippers were added.}
\label{fig:g1_hardware_nv}
\end{figure}

%% file: tables/g1-hardware-nv.tex
\begin{table}[H]
\centering
\small
\setlength{\tabcolsep}{5pt}
\renewcommand{\arraystretch}{1.12}
\caption{\textbf{Sensor and state interfaces.} Nominal specifications and
configured publication rates. Lidar angles use the manufacturer's orientation.}
\label{tab:g1_hardware_nv}
\begin{tabularx}{\linewidth}{@{}>{\raggedright\arraybackslash}p{0.18\linewidth}>{\raggedright\arraybackslash}X >{\raggedright\arraybackslash}X@{}}
\toprule
Component & Native input / specification & Configured output \\
\midrule
Insta360 X5 & USB panorama: $2880\!\times\!1440$; 30\,fps~\citep{insta360x5webcam}
  & 6\,Hz JPEG; 180\degree\ yaw correction \\
Livox Mid-360 & 200,000 points/s; field of view: 360\degree\ H, $-7\degree$ to $52\degree$ V~\citep{livoxmid360specs}
  & SLAM bridge: 10\,Hz; at most 20,000 points/batch \\
Body state & 29 joint positions/velocities; IMU, odometry and mode
  & 20\,Hz \\
Head OAK camera & Calibrated RGB and stereo depth
  & RGB/depth observations; camera-to-body calibration \\
YAM grippers\newline (optional) & Two parallel grippers with fin-ray fingers
  & Opening/closing commands; jaw width and controller state \\
\bottomrule
\end{tabularx}
\end{table}

%% file: tables/g1-drive-nv.tex
\begin{table}[H]
\centering
\small
\setlength{\tabcolsep}{5pt}
\renewcommand{\arraystretch}{1.1}
\caption{\textbf{Navigation settings and command limits.} The slow-walk controller
limits the speed, duration, and size of individual commands and path segments.}
\label{tab:g1_drive_nv}
\begin{tabularx}{\linewidth}{@{}>{\raggedright\arraybackslash}X l >{\raggedright\arraybackslash}X l@{}}
\toprule
Parameter & Value & Parameter & Value \\
\midrule
Command publication & 10\,Hz & Path lookahead & 0.40\,m \\
Translation speed bounds & 0.12--0.35\,m/s & Braking lead time & 0.60\,s \\
Near-goal speed ($\le0.5$\,m) & 0.25\,m/s & Velocity-command expiry & 2.0\,s \\
Arrival tolerances & 0.10\,m; 4\degree & Active-job heartbeat timeout & 3.0\,s \\
Move / turn command limits & 2.0\,m; 180\degree & Path chunk limits & 6.0\,m; 16 points \\
Move-job timeout & 25\,s & Path-job timeout & 45\,s \\
\bottomrule
\end{tabularx}
\end{table}